%% file: main.tex
\documentclass[conference]{IEEEtran}
\IEEEoverridecommandlockouts
\usepackage{amsmath,amssymb,amsfonts}
\usepackage{graphicx}
\usepackage{textcomp}
\usepackage{xcolor}
\usepackage{svg}
\def\BibTeX{{\rm B\kern-.05em{\sc i\kern-.025em b}\kern-.08em
    T\kern-.1667em\lower.7ex\hbox{E}\kern-.125emX}}

\include{include}
\include{acronym}

\renewcommand{\baselinestretch}{0.95}

\newcommand{\luca}[1]{{\color{black}#1}}
\newcommand{\gapflow}[1]{{\color{black}#1}}

\makeatletter
\def\ps@IEEEtitlepagestyle{%
  \def\@oddfoot{\mycopyrightnotice}%
  \def\@oddhead{\hbox{}\@IEEEheaderstyle\leftmark\hfil\thepage}\relax
  \def\@evenhead{\@IEEEheaderstyle\thepage\hfil\leftmark\hbox{}}\relax
  \def\@evenfoot{}%
}
\def\mycopyrightnotice{%
  \begin{minipage}{\textwidth}
  \centering \scriptsize
  \copyright 2021 IEEE.  Personal use of this material is permitted.  Permission from IEEE must be obtained for all other uses, in any current or future media, including reprinting/republishing this material for advertising or promotional purposes, creating new collective works, for resale or redistribution to servers or lists, or reuse of any copyrighted component of this work in other works.
  \end{minipage}
}
\makeatother

\begin{document}

\title{
%Temporal Convolutional Neural Networks on Parallel Ultra-Low-Power Platform for Cardiac Arrhythmia Classification
%An Accurate 
% energy-efficient
%A Temporal Convolutional Network for Wearable Cardiac Arrhythmia Classification
%ECG-TCN: A Temporal Convolutional Network for Wearable Cardiac Arrhythmia Classification
ECG-TCN: Wearable Cardiac Arrhythmia Detection with a Temporal Convolutional Network
}

%\author{
%\IEEEauthorblockN{Authors removed for double-blind review.\vspace{-1cm}}
%\IEEEauthorblockA{}
%}

 \author{\IEEEauthorblockN{
     Thorir Mar Ingolfsson\IEEEauthorrefmark{1}, 
     Xiaying Wang\IEEEauthorrefmark{1}, 
     Michael Hersche\IEEEauthorrefmark{1}, 
     Alessio Burrello\IEEEauthorrefmark{3},
     Lukas Cavigelli\IEEEauthorrefmark{2},
     Luca Benini\IEEEauthorrefmark{1}\IEEEauthorrefmark{3}}
     \IEEEauthorblockA{\\\IEEEauthorrefmark{1}ETH Zürich, D-ITET, Switzerland \hspace{3.2mm}\IEEEauthorrefmark{3}University of Bologna, DEI, Italy \hspace{3.2mm}
     \IEEEauthorrefmark{2}Huawei Technologies, Zurich RC, Switzerland\vspace{-0.5cm}}
     \thanks{Corresponding emails: \{thoriri, xiaywang\}@iis.ee.ethz.ch}
     }

\maketitle

\begin{abstract}
\luca{
Personalized ubiquitous healthcare solutions require energy-efficient wearable platforms that provide an accurate classification of bio-signals while consuming low average power for long-term battery-operated use. 
}
Single lead electrocardiogram (ECG) signals provide the ability to detect, classify, and even predict cardiac arrhythmia.
%
%Real-time health monitoring allows users to live their life to the fullest, knowing their wearable device will notify them of any irregularities.
%
%Single lead electrocardiogram (ECG) signals provide the ability to detect, classify, and even predict cardiac arrhythmia.
%
%Personalized on-demand healthcare solutions require energy-efficient wearable platforms that provide accurate classifications while consuming low average power to be battery-operated for long-term use.
%
In this paper we propose a novel temporal convolutional network (TCN) that achieves high accuracy while still being feasible for wearable platform use.
Experimental results on the ECG5000 dataset show that the TCN has a similar accuracy (94.2\%) score as the state-of-the-art (SoA) network while achieving an improvement of 16.5\% in the balanced accuracy score. This accurate classification is done with 27$\times$ fewer parameters and 37$\times$ less multiply-accumulate operations.
We test our implementation on two publicly available platforms, the STM32L475, which is based on ARM Cortex M4F, and the GreenWaves Technologies GAP8 on the GAPuino board, based on 1+8 RISC-V CV32E40P cores.
\luca{
Measurements show that the GAP8 implementation respects the real-time constraints while consuming 0.10\,mJ per inference. With 9.91\,GMAC/s/W, it is 23.0$\times$ more energy-efficient and 46.85$\times$ faster than an implementation on the ARM Cortex M4F (0.43\,GMAC/s/W).
Overall, we obtain 8.1\% higher accuracy while consuming 19.6$\times$ less energy and being 35.1$\times$ faster compared to a previous SoA embedded implementation.
}
\end{abstract}


\input{introduction}
\input{methods}
\input{results}

\section{Conclusion}\label{ch:conclusion}
This paper presents a novel, lightweight TCN architecture for cardiac arrhythmia classification. The network achieves similar accuracy results as the SoA on the ECG5000 dataset while improving the balanced accuracy score by 16.5\%. The TCN is then further quantized to INT-8 using two different quantization tools and implemented on two publicly available platforms, STM32L475 and GAP8. The INT-8 quantized TCN implementation on the GAP8 platform with the NEMO quantization scheme and DORY tool for deployment shows superior results, being 23.0$\times$ more energy-efficient than the implementation on the ARM Cortex M4F. 
\luca{
The implementation has a throughput of 381.58\,MMAC/s, consumes 0.10\,mJ per inference, and takes only 2.7\,ms to output a prediction. Integrating an ADC such as~\cite{6243792} with GAP8 would allow us to build an ECG smartpatch consuming 382.23\,$\mu$W on average. Assuming a real-world scenario with one inference per second, i.e an average heartbeat rate of 60bpm, the system would last two years on a CR2450 coin cell battery.
}
\section*{Acknowledgment}
\vspace{-0.1cm}
This project was supported in part by the Swiss National Science Foundation (Project PEDESITE) under grant agreement CRSII5\_193813 and in part by the Swiss Data Science Center PhD Fellowship under grant ID P18-04.
%\tiny
\bibliographystyle{IEEEtran}
\bibliography{bibliography,ref_michael_mendeley}

\end{document}

%% file: include.tex
\usepackage{comment}

\usepackage{nohyperref}

\definecolor{matplotlib0}{HTML}{1f77b4}
\definecolor{matplotlib1}{HTML}{d62728}
\definecolor{matplotlib2}{HTML}{2ca02c}
\definecolor{matplotlib3}{HTML}{ff7f0e}
\definecolor{matplotlib4}{HTML}{9467bd}
\definecolor{matplotlib5}{HTML}{8c564b}
\definecolor{matplotlib6}{HTML}{e377c2}
\definecolor{matplotlib7}{HTML}{7f7f7f}
\definecolor{matplotlib8}{HTML}{bcbd22}
\definecolor{matplotlib9}{HTML}{17becf}

\usepackage{mathtools} % \abs, ...

\usepackage{booktabs}
\usepackage{multirow}
\usepackage{colortbl}
\usepackage{tablefootnote}
\usepackage{threeparttable}

\usepackage[acronym, style=super, nonumberlist]{glossaries}

\usepackage{pgfplots}
\usepgfplotslibrary{fillbetween}
\usepgfplotslibrary{colormaps}
\pgfplotsset{compat=1.16}

\pgfplotscreateplotcyclelist{matplotlib}{
  {matplotlib0},
  {matplotlib1},
  {matplotlib2},
  {matplotlib3},
  {matplotlib4},
  {matplotlib5},
  {matplotlib6},
  {matplotlib7},
  {matplotlib8},
  {matplotlib9}
}

\pgfplotsset{every axis/.append style={
    cycle list name=matplotlib,
}}

\usepackage{listings}

\definecolor{code_default}{HTML}{000000}
\definecolor{code_keyword}{HTML}{AC4142}
\definecolor{code_identifier}{HTML}{D28445}

\lstdefinelanguage{RISCV}{
  sensitive=false,
  morecomment=[l]{//},
  alsoletter={.},
  morekeywords=[1]{
    lp.setup, mv, lw, p.lw, sw, p.sw, pv.sdotsp.b, pv.shuffle2.b, p.subNR, p.addNR
  },
  morekeywords=[2]{
    zero, ra, sp, gp, tp, t0, t1, t2, t3, t4, t5, t6, s0, s1, a0, a1, a2, a3, a4, a5, a6, a7, a8, a9, a10, a11,
  },
  morestring=[b]",
  morestring=[b]',
}[strings, comments, keywords]

\lstdefinestyle{RISCV_STYLE}{
  language=RISCV,
  numbers=none,
  basicstyle=\scriptsize\ttfamily\color{code_default},
  keywordstyle=[1]\color{matplotlib0},
  keywordstyle=[2]\color{matplotlib1},
  float,
  captionpos=b,
  belowskip=-0.5cm
}

\usepackage{algorithm}
\usepackage{algpseudocode}
\usepackage{float}
\newfloat{algorithm}{t}{top}

%% file: acronym.tex
\newacronym{simd}{SIMD}{Single Instruction, Multiple Data}
\newacronym{elu}{ELU}{Exponential Linear Unit}
\newacronym{relu}{ReLU}{Rectified Linear Unit}
\newacronym{rpr}{RPR}{Random Partition Relaxation}
\newacronym{mac}{MAC}{Multiply Accumulate}
\newacronym{dma}{DMA}{Direct Memory Access}
\newacronym{bmi}{BMI}{Brain--Machine Interface}
\newacronym{bci}{BCI}{Brain--Computer Interface}
\newacronym{smr}{SMR}{Sensory Motor Rythms}
\newacronym{eeg}{EEG}{Electroencephalography}
\newacronym{svm}{SVM}{Support Vector Machine}
\newacronym{svd}{SVD}{Singular Value Decomposition}
\newacronym{evd}{EVD}{Eigendecomposition}
\newacronym{iir}{IIR}{Infinite Impulse Response}
\newacronym{fir}{FIR}{Finite Impulse Response}
\newacronym{fc}{FC}{Fabric Controller}
\newacronym{nn}{NN}{Neural Network}
\newacronym{mrc}{MRC}{Multiscale Riemannian Classifier}
\newacronym{flop}{FLOP}{Floating Point Operation}
\newacronym{sos}{SOS}{Second-Order Section}
\newacronym{ipc}{IPC}{Instructions per Cycle}
\newacronym{tcdm}{TCDM}{Tightly Coupled Data Memory}
\newacronym{fpu}{FPU}{Floating Point Unit}
\newacronym{fma}{FMA}{Fused Multiply Add}
\newacronym{alu}{ALU}{Arithmetic Logic Unit}
\newacronym{dsp}{DSP}{Digital Signal Processing}
\newacronym{gpu}{GPU}{Graphics Processing Unit}
\newacronym{soc}{SoC}{System-on-Chip}
\newacronym{mi}{MI}{Motor-Imagery}
\newacronym{csp}{CSP}{Commmon Spatial Patterns}
\newacronym{fbcsp}{FBCSP}{Filter-Bank \acrlong{csp}}
\newacronym{pulp}{PULP}{parallel ultra-low power}
\newacronym{soa}{SoA}{state-of-the-art}
\newacronym{bn}{BN}{Batch Normalization}
\newacronym{isa}{ISA}{Instruction Set Architecture}
\newacronym{ecg}{ECG}{Electrocardiogram}
\newacronym{mcu}{MCU}{microcontroller}
\newacronym{rnn}{RNN}{recurrent neural network}
\newacronym{cnn}{CNN}{convolutional neural network}
\newacronym{tcn}{TCN}{temporal convolutional network}

%% file: introduction.tex
\section{Introduction}\label{ch:introduction}
%% abstract + intro 1 page
% wearables w/ ECG for health
Wearable devices are becoming increasingly popular for health monitoring. They enable to continuously track vital parameters, assist their users to live healthy, and potentially detect health disorders~\cite{wearablesandmedical}. 
For example, hard-to-diagnose diseases such as sleep apnoea or intermittent cardiac arrhythmia can be identified more easily, reliably, and earlier. Particularly high-risk patients can benefit from always-on cardiovascular monitoring where the early detection of cardiac arrhythmia, or a potential infarction, increase the chances of full recovery; and thus, can be life saving. 
\gls{ecg} signals are commonly used to identify different types of heart diseases and particularly detect acute disease states. Traditionally, the acquired signal is either recorded but not analyzed in case of diagnostic devices, processed by a larger non-wearable device available in medical care facilities, or in the rare cases of wearable monitoring transmitted to a remote cloud where it is processed and analyzed. While the first have clear limits in capability and usability, the latter exhibits a high energy consumption for the data transmission and gives rise to privacy concerns associated with personal healthcare data. 

% platforms, tools, flows
With the recent technological advancements in low-power microprocessors, e.g., \glspl{mcu} based on ARM Cortex M family or RISC-V \gls{pulp} platform~\cite{8445101}, smart edge computing is becoming the new compute paradigm to overcome these limitations. Here, the sensor data is directly processed locally on the smart wearable device, right where it is collected~\cite{infiniwolf}.
On the data analysis side, deep learning has shown impressive performance leaps across many fields. For time series classification tasks, \glspl{cnn} and \glspl{rnn} are now setting the \gls{soa} accuracy while simplifying the overall engineering process by omitting the development of application-specific engineered features \cite{karim_lstm_2018}. 
With the increasing popularity of deep learning, many frameworks have become available to efficiently deploy neural networks not only in the cloud, but also to various edge devices. However, current comparison efforts on their effectiveness and usability have been limited, often hindered by device-specific deployment flows.

However, many deep learning models are also known to require notoriously large amounts of resources in terms of compute, power, and memory, which remain out of reach for low-power \glspl{mcu}. Searching for smaller networks that fit the constraints of the platforms, engineers must accept a lower accuracy~\cite{faraone2020aicas}. Hence, active research is on-going to design resource-efficient models that are compact. A viable option are \glspl{tcn}~\cite{bai_empirical_2018,9159637} which have shown to achieve high accuracy in biosignal classification even with small model size and limited demand for resources~\cite{9283028}. 

\begin{figure*}[ht!]
    \fontsize{6}{7}\selectfont
    \centering
    %\def\svgwidth{\columnwidth}
    %\includesvg[width=\textwidth]{figures/tcn_ecg}
    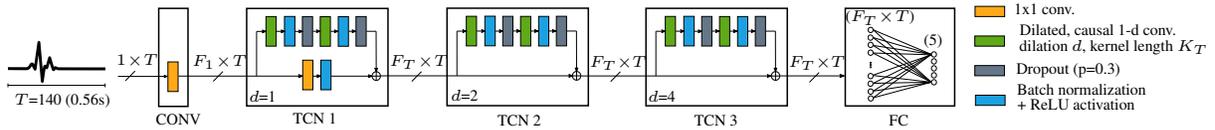
    \caption{TCN architecture for arrhythmia detection processing 140 ECG samples (0.56\,s). The first convolutional block (CONV) has $F_1=2$ filters. TCN blocks 1--3 use a same number of filters $F_T=11$, the same kernel length $K_T=11$, and a variable dilation $d\in\{1,\dots,4\}$. 
    }
    \label{fig:resblocks}
\end{figure*}
% NOTE: we maybe should probably implement the TCN also with GWT Autotiler and nntool??

% Most of the current deep learning models for time-series inference demand high resources in terms of compute, power, and memory. 
%
% cite Farone (AICAS'20), Gap8
% Maybe give an example of Farone, however, they achieve relatively low accuracy due to the small network. 
%
% Therefore, there is need for resource-efficient models that are compact yet accurate. 
% 
% recent developments in TinyML. many frameworks, very active research area. Which frameworks are better?
% A viable option are \glspl{tcn}~\cite{bai_empirical_2018} which have shown to achieve high accuracy in biosignal classification tasks such as EEG~\cite{9283028}. 
% 
% Some details about TCNs in general. 

% 

In this paper, we propose a compact TCN for accurate time-series classification on embedded devices, demonstrated with the deployment and characterization of an ECG-based heartbeat classifier on multiple embedded platforms. 
The main contributions are: 
\begin{itemize}
    \item We propose a novel TCN for accurate heartbeat classification. 
    On the ECG5000 dataset, our proposed model achieves similar accuracy as the SoA LSTM-FCN network~\cite{karim_lstm_2018} (94.2\% vs. 94.1\%) and higher balanced accuracy (89.0\% vs 72.5\%), while requiring an order-of-magnitude lower number of parameters (15\,k vs 405\,k) and multiply-accumulate operations (1\,MMACs vs 37\,MMACs). 
    %\item We quantize the TCN to INT-8 and deploy the model on an ARM Cortex M4F using either the TFLite or the \textsc{Cube.AI} deployment framework, and on a RISC-V, PULP-based GreenWaves Technologies (GWT) GAP8~\cite{8445101} \gls{mcu} using NEMO and DORY tools~\cite{burrello2020dory}. 
    \gapflow{\item We quantize to INT-8 and deploy the model on an ARM Cortex M4F using both TFLite and \textsc{Cube.AI} deployment framework, and on a RISC-V PULP-based GreenWaves Technologies (GWT) GAP8~\cite{8445101} \gls{mcu} using both the GWT-proprietary and closed-source GAPflow and the academic, free open-source NEMO/DORY tools~\cite{burrello2020dory}.}
    \item We report extensive throughput and power measurements, comparing various deployment frameworks and target platforms.
    The proposed TCN achieves highest energy efficiency (9.91\,GMAC/s/W) when being deployed on a GAP8 using NEMO and DORY. 
    Compared to a prior art CNN+GRU deployed on a Cortex M4F using CMSIS-NN~\cite{faraone2020aicas}, our deployed TCN consumes 19.6$\times$ less energy (i.e., 6.0$\times$ more energy-efficient) and is 35.1$\times$ faster, while being 8.1\% more accurate. 
\end{itemize}
We open-source release our codes\footnote{https://github.com/pulp-platform/ECG-TCN}, including the enhancements made to the open-source DORY tool.

%%COMMENTS FROM THORIR
%The reference paper we are working with is here: https://ieeexplore.ieee.org/stamp/stamp.jsp?tp=&arnumber=9073950&tag=1 it's done on another dataset but it's still a single lead ECG dataset so that's incommon. They use an ARM M4 and use CMSIS-NN to get the network over to the MCU.

%Work done on this ECG5000 dataset includes: LSTM-FCN, which can be found here: https://ieeexplore.ieee.org/stamp/stamp.jsp?tp=&arnumber=8141873 they do benchamarking with the LSTM-FCN and ALSTM-FCN on the University of California Riverside (UCR) benchmark dataset and ECG5000 is a subset of datasets within this UCR dataset.
%Note that we actually reproduced their results and those reproductions we are showing in our tables.
%We then also split the LSTM-FCN parts and that's the LSTM and CNN parts that we also reproduced and are including in our tables.

%Then we also have this here work: https://dl.acm.org/doi/10.1145/3274783.3275208 which is the 2 page paper that also did work on ECG5000 but we don't have enough info to include stuff on MACs and Parameters in our comparison here.

%Also have a look at the paragraph in my master thesis on related works into ECG5000, there are some other links there to other methods.

%%% Local Variables:
%%% mode: latex
%%% TeX-master: "report"
%%% End:

%% file: figures/tcn_ecg.pdf_tex
%% Creator: Inkscape inkscape 0.92.4, www.inkscape.org
%% PDF/EPS/PS + LaTeX output extension by Johan Engelen, 2010
%% Accompanies image file 'figures/tcn_ecg.pdf' (pdf, eps, ps)
%%
%% To include the image in your LaTeX document, write
%%   \input{<filename>.pdf_tex}
%%  instead of
%%   \includegraphics{<filename>.pdf}
%% To scale the image, write
%%   \def\svgwidth{<desired width>}
%%   \input{<filename>.pdf_tex}
%%  instead of
%%   \includegraphics[width=<desired width>]{<filename>.pdf}
%%
%% Images with a different path to the parent latex file can
%% be accessed with the `import' package (which may need to be
%% installed) using
%%   \usepackage{import}
%% in the preamble, and then including the image with
%%   \import{<path to file>}{<filename>.pdf_tex}
%% Alternatively, one can specify
%%   \graphicspath{{<path to file>/}}
%% 
%% For more information, please see info/svg-inkscape on CTAN:
%%   http://tug.ctan.org/tex-archive/info/svg-inkscape
%%
\begingroup%
  \makeatletter%
  \providecommand\color[2][]{%
    \errmessage{(Inkscape) Color is used for the text in Inkscape, but the package 'color.sty' is not loaded}%
    \renewcommand\color[2][]{}%
  }%
  \providecommand\transparent[1]{%
    \errmessage{(Inkscape) Transparency is used (non-zero) for the text in Inkscape, but the package 'transparent.sty' is not loaded}%
    \renewcommand\transparent[1]{}%
  }%
  \providecommand\rotatebox[2]{#2}%
  \newcommand*\fsize{\dimexpr\f@size pt\relax}%
  \newcommand*\lineheight[1]{\fontsize{\fsize}{#1\fsize}\selectfont}%
  \ifx\svgwidth\undefined%
    \setlength{\unitlength}{457.34071727bp}%
    \ifx\svgscale\undefined%
      \relax%
    \else%
      \setlength{\unitlength}{\unitlength * \real{\svgscale}}%
    \fi%
  \else%
    \setlength{\unitlength}{\svgwidth}%
  \fi%
  \global\let\svgwidth\undefined%
  \global\let\svgscale\undefined%
  \makeatother%
  \begin{picture}(1,0.09844244)%
    \lineheight{1}%
    \setlength\tabcolsep{0pt}%
    \put(-0.27218191,-0.19711617){\color[rgb]{0,0,0}\makebox(0,0)[lt]{\begin{minipage}{1.44312542\unitlength}\raggedright \end{minipage}}}%
    \put(0,0){\includegraphics[width=\unitlength,page=1]{figures/tcn_ecg.pdf}}%
    \put(0.77624884,0.06616019){\color[rgb]{0,0,0}\makebox(0,0)[t]{\lineheight{1.25}\smash{\begin{tabular}[t]{c}(5)\end{tabular}}}}%
    \put(0,0){\includegraphics[width=\unitlength,page=2]{figures/tcn_ecg.pdf}}%
    \put(0.7325,0.0858881){\color[rgb]{0,0,0}\makebox(0,0)[t]{\lineheight{1.25}\smash{\begin{tabular}[t]{c}($F_T\!\times\!T$)\end{tabular}}}}%
    \put(0,0){\includegraphics[width=\unitlength,page=3]{figures/tcn_ecg.pdf}}%
    \put(0.88755604,0.04257937){\color[rgb]{0,0,0}\makebox(0,0)[t]{\lineheight{1.25}\smash{\begin{tabular}[t]{c}Dropout (p=0.3)\end{tabular}}}}%
    \put(0.91207341,0.07417832){\color[rgb]{0,0,0}\makebox(0,0)[t]{\lineheight{1.25}\smash{\begin{tabular}[t]{c}Dilated, causal 1-d conv.\end{tabular}}}}%
    \put(0.92499932,0.06148786){\color[rgb]{0,0,0}\makebox(0,0)[t]{\lineheight{1.25}\smash{\begin{tabular}[t]{c} dilation $d$, kernel length $K_T$  \end{tabular}}}}%
    \put(0.89854137,0.02489555){\color[rgb]{0,0,0}\makebox(0,0)[t]{\lineheight{1.25}\smash{\begin{tabular}[t]{c}Batch normalization\end{tabular}}}}%
    \put(0.89409001,0.01282612){\color[rgb]{0,0,0}\makebox(0,0)[t]{\lineheight{1.25}\smash{\begin{tabular}[t]{c}+ ReLU activation\end{tabular}}}}%
    \put(0,0){\includegraphics[width=\unitlength,page=4]{figures/tcn_ecg.pdf}}%
    \put(0.86805398,0.0916637){\color[rgb]{0,0,0}\makebox(0,0)[t]{\lineheight{1.25}\smash{\begin{tabular}[t]{c}1x1 conv. \end{tabular}}}}%
    \put(0.15436196,0.00081983){\color[rgb]{0,0,0}\makebox(0,0)[t]{\lineheight{1.25}\smash{\begin{tabular}[t]{c}CONV\end{tabular}}}}%
    \put(0.26666307,0.00144374){\color[rgb]{0,0,0}\makebox(0,0)[t]{\lineheight{1.25}\smash{\begin{tabular}[t]{c}TCN 1\end{tabular}}}}%
    \put(0.43522958,0.00020159){\color[rgb]{0,0,0}\makebox(0,0)[t]{\lineheight{1.25}\smash{\begin{tabular}[t]{c}TCN 2\end{tabular}}}}%
    \put(0.59729573,0.00018937){\color[rgb]{0,0,0}\makebox(0,0)[t]{\lineheight{1.25}\smash{\begin{tabular}[t]{c}TCN 3\end{tabular}}}}%
    \put(0.74782898,0.00020159){\color[rgb]{0,0,0}\makebox(0,0)[t]{\lineheight{1.25}\smash{\begin{tabular}[t]{c}FC\end{tabular}}}}%
    \put(0.05863734,0.0174538){\color[rgb]{0,0,0}\makebox(0,0)[t]{\lineheight{1.25}\smash{\begin{tabular}[t]{c}$T$=140 (0.56s)\end{tabular}}}}%
    \put(0,0){\includegraphics[width=\unitlength,page=5]{figures/tcn_ecg.pdf}}%
    \put(0.16584434,0.04931415){\color[rgb]{0,0,0}\makebox(0,0)[lt]{\lineheight{1.25}\smash{\begin{tabular}[t]{l}$F_1\!\times\!T$\end{tabular}}}}%
    \put(0,0){\includegraphics[width=\unitlength,page=6]{figures/tcn_ecg.pdf}}%
    \put(0.32861383,0.04928458){\color[rgb]{0,0,0}\makebox(0,0)[lt]{\lineheight{1.25}\smash{\begin{tabular}[t]{l}$F_T\! \times\!T$\end{tabular}}}}%
    \put(0.49343512,0.04841549){\color[rgb]{0,0,0}\makebox(0,0)[lt]{\lineheight{1.25}\smash{\begin{tabular}[t]{l}$F_T $$\times$$T$\end{tabular}}}}%
    \put(0.65718069,0.04841549){\color[rgb]{0,0,0}\makebox(0,0)[lt]{\lineheight{1.25}\smash{\begin{tabular}[t]{l}$F_T$$\times$$T$\end{tabular}}}}%
    \put(0,0){\includegraphics[width=\unitlength,page=7]{figures/tcn_ecg.pdf}}%
    \put(0.10330489,0.04974169){\color[rgb]{0,0,0}\makebox(0,0)[lt]{\lineheight{1.25}\smash{\begin{tabular}[t]{l}$1\!\times\!T$\end{tabular}}}}%
    \put(0.21250788,0.01920226){\color[rgb]{0,0,0}\makebox(0,0)[lt]{\lineheight{1.25}\smash{\begin{tabular}[t]{l}$d$=1\end{tabular}}}}%
    \put(0.3788061,0.02029471){\color[rgb]{0,0,0}\makebox(0,0)[lt]{\lineheight{1.25}\smash{\begin{tabular}[t]{l}$d$=2\end{tabular}}}}%
    \put(0.54141939,0.01942501){\color[rgb]{0,0,0}\makebox(0,0)[lt]{\lineheight{1.25}\smash{\begin{tabular}[t]{l}$d$=4\end{tabular}}}}%
    \put(0,0){\includegraphics[width=\unitlength,page=8]{figures/tcn_ecg.pdf}}%
  \end{picture}%
\endgroup%

%% file: methods.tex
\section{Methods}\label{ch:background}

\subsection{Temporal Convolutional Networks}\label{ch:background:tcn}
The network proposed in this paper is commonly referred to as a temporal convolutional network (TCN) and is depicted in Fig.~\ref{fig:resblocks}. Before passing the ECG signal through the TCN blocks, we expand the channels of the signal by passing them through a 1x1 convolutional layer.
The residual block of our TCN consists of two layers of dilated convolutions, with batch normalization, non-linearity, and a dropout layer in-between the convolutions. 
Even though TCNs feature only 1D convolutions, they can still process 2D feature maps by considering the second channel dimension as the depth dimension.
The residual skip connection adds the input to the output feature map, with the check that if the depth of the input and output is different, a $1\times 1$ convolution is inserted.
%to ensure the tensors are of the same shape. 
%
See Fig.~\ref{fig:resblocks} for an illustration of a residual block and the stacking of three residual blocks together.

By stacking residual blocks, the receptive field size ($\mathrm{RFS}$) increases exponentially with each residual block, as the dilation in each subsequent block grows exponentially larger by the factor $d = 2^{L-1}$, where $L$ is the number of the residual block. 
%
%This gives us a nifty formula to calculate the receptive field size (RFS) of the TCN:
The RFS of the TCN is then determined by
\begin{align*}
    \mathrm{RFS} = 1 + 2 \cdot (\textit{$K_T$} - 1)\cdot(2^{\textit{L}} - 1),
\end{align*}
where $K_T$ is the kernel size.
%{\color{red} the dilation should be in this formula. let's put somewhere in the text that $d = 2^L$}

%The residual block of our TCN slightly differs from the initially introduced form~\cite{bai_empirical_2018}: 
The residual block of our TCN differs from the one initially proposed by~\cite{bai_empirical_2018}:
\begin{itemize}
    \item Batch normalization is used between convolutions instead of weight normalization as batch normalization has been shown to provide higher accuracy than weight normalization on various large scale networks~\cite{gitman_comparison_2017}.
    \item We use batch normalization on the residual branch wherever a $1\times 1$ convolution is inserted.
    \item Instead of spatial dropout, normal dropout is used. As the TCN is applied after various convolutions, the adjacent samples within feature maps are not strongly correlated, hence it is beneficial to drop individual elements instead of entire 1D feature maps to regularize the activations.
\end{itemize}

\subsection{Training Procedure}
The model was developed, trained, and tested on an Nvidia GTX 1080 Ti GPU using both a TensorFlow (TF) and a PyTorch environment, to adapt to the different deployment frameworks.
%
%The networks were developed jointly with TensorFlow and PyTorch.
%
When training the proposed TCN, we use categorical cross-entropy loss and uniformly initialize filter kernels with the procedure introduced in~\cite{he_deep_2016}.
The TCN model is trained on batches of size 30 using an Adam optimizer with a learning rate of 0.001 for 20 epochs.
The final network is chosen via cross-validation on the training set and the final accuracy is reported on the test set.
\subsection{Hardware Deployment}
We target two platforms to be able to compare performance and find the best-suited microprocessor.
\subsubsection{GAP8}
The first platform is GWT GAPuino with GAP8, a RISC-V-based PULP platform with two compute domains: a fabric controller with one CV32E40P core for control tasks and 512\,kB L2 memory extendable via a HyperBus interface and a cluster domain with 8 CV32E40P cores for parallel computation of highly demanding workloads and 80\,kB directly accessible L1 memory. The frequency ranges from 32\,kHz to 250\,MHz. 
\gapflow{
We compare two deployment frameworks:

\begin{itemize}
    \item \emph{GAPflow}: 
    We quantized our network using TFLite and deployed it using GAPflow. 
    It comprises of NNTool, which parses and prepares TFLite models, and AutoTiler, which converts the parsed model into optimized C code that can be executed efficiently in parallel by the 8 cores in the cluster.
    Since dilated causal convolutions are currently not supported, we replaced the dilated causal convolutional blocks as seen in Fig.~\ref{fig:resblocks} in the already trained model with the functionally identical sequence of a padding layer which pads the signal on the left side with $d \cdot (K_T-1)$ zeros (causal padding), followed by a 1D convolution layer with a zero-stuffed weight kernel to imitate dilation. Moreover, the batch normalization layers are also removed without affecting the accuracy.
    \item \emph{NEMO/DORY}: We used NEMO~\cite{nemo} to quantize our network to INT-8 and deployed it using DORY~\cite{burrello2020dory}. The open-source tool automatically generates C code to manage the two-level memory, L1 and L2, and execute convolutional kernels while maximizing the data locality. Similar to AutoTiler, DORY uses tiling and double-buffering to hide memory movements during kernel computation, acting as a software cache and minimizing the memory latency. While DORY employs PULP-NN kernels~\cite{Garofalo2019PULP-NN} optimized for 2D convolutions, in our work, we extended DORY with a novel, optimized 1D kernel version of the original 2D kernels, inserting a variable dilation parameter and optimizing the parallelization together with the internal loop execution. These kernels plug into the DORY templates for automatic network code generation. Further, we adapted the DORY memory allocation to support convolutions on two parallel branches of the network, as present in our residual blocks.
\end{itemize}
}

% % NEMO DORY
% To efficiently deploy our network and exploit the 8-cores parallel cluster, we rely on the DORY tool~\cite{burrello2020dory}. %, which outperforms GWT Autotiler for classical convolutions.
% %
% The tool automatically generates C code to manage the two-level memory, L1 and L2, and execute convolutional kernels while maximizing the data locality.
% %

% DORY uses tiling and double-buffering to hide memory movements during kernel computation, acting as a software cache and minimizing the memory latency.
% %
% While DORY employs PULP-NN kernels~\cite{Garofalo2019PULP-NN}, in our work, we developed a novel, optimized 1-D kernel version of the original 2D kernels, inserting a variable dilation parameter and optimizing the parallelization together with the internal loop execution.
% %
% These kernels plug into the DORY templates for automatic network code generation.
% %
% Further, we adapted the DORY memory allocation to support convolutions on two parallel branches of the network, as present in our residual blocks.
% %
% We also used NEMO~\cite{nemo} to quantize our network to INT-8, as required by the DORY tool.

\subsubsection{ARM Cortex M4F}
The second platform is the B-L475E-IOT01A STM32L4 Discovery kit, which is based on an ARM Cortex M4F core with 1\,MB of flash memory, 128\,kB of SRAM, and clocked at 80\,MHz.
We explore three different ways of deploying networks on the Cortex M4F and discuss their advantages and disadvantages:
\begin{itemize}
    \item \emph{TFLite Micro:} It provides a library and a tool flow to convert trained TF models to MCU-compatible code with some limitations for non-frequently used layers. We built TFLite Micro from source with flags for the Cortex M4F architecture and indicating that TFLite should use the optimized CMSIS-NN kernels provided by ARM. The model is built by calling layers from the resulting TFLite Micro library. With this method, it is possible to run quantized INT-8 TFLite models on the ARM Cortex M4F.
    \item \emph{\textsc{Cube.AI} with TFLite:} The use of \textsc{Cube.AI} allows the automatic conversion of pre-trained neural networks and integration of generated optimized library into project code. The latest version of \textsc{Cube.AI} takes a quantized INT-8 TFLite model and automatically parses the network graph, and generates an optimized code to run the network on inputs. It also can take a floating-point model made with Keras as an input and generates code to run the model on the MCU.
    \item \emph{\textsc{Cube.AI} with Keras:} This method is similar to the previous \textsc{Cube.AI}-based flow but starts from a Keras model instead. However, not all the layers are quantized to an INT-8 representation in this mode.
\end{itemize}
\gapflow{
As for GAPflow, manual padding is added to the dilated causal convolutions for the deployments on ARM since this operation is currently supported by neither \textsc{Cube.AI} nor TFLite.
}
%\subsection{Network Architecture Differences}
% The TCN depicted in Fig.~\ref{fig:resblocks} needs to be adapted slightly before being implemented on the Cortex M4F as dilated causal convolutions are currently not supported by \textsc{Cube.AI} and TFLite. We replaced the dilated causal convolutional blocks as seen in Fig.~\ref{fig:resblocks} in the already trained model with the functionally identical sequence of a padding layer which pads the signal on the left side with $d \cdot (K_T-1)$ zeros (causal padding), followed by a 1D convolution layer with a zero-stuffed weight kernel to imitate dilation.

\subsection{Performance Metrics}
We evaluate the models according to the classification accuracy, which is the ratio between correctly classified trials and the total number of trials in the test set. 
The ECG5000 dataset we use in this paper is highly imbalanced, having only two samples of one of the classes in the whole training set. Therefore, we also consider the balanced accuracy score defined as $(\textrm{Sensitivity}+\textrm{Specificity})/2$.
% \begin{align*}
%     \frac{\textnormal{Sensitivity}+\textnormal{Specificity}}{2}.
% \end{align*}
%
To measure how computationally intensive each network is we also compute the number of MACs for inference.

% To circumvent this problem we mimicked the effects of the dilated causal convolutions by using operations supported by TFLite Micro. Therefore instead of the dilated causal convolutional blocks as seen in Fig.~\ref{fig:resblocks} we have a padding layer which pads the signal on the left side by $d \cdot (K_T-1)$ this is called causal padding. Then to implement the dilation of the convolution the kernel size was increased to  $d \cdot (K_T-1) + 1$, then at the correct places, zero kernel weights were fixed into the network, so to force it to behave like a dilated convolution.

%%% Local Variables:
%%% mode: latex
%%% TeX-master: "report"
%%% End:

%% file: results.tex
\section{Experimental Results}\label{ch:results}
%\subsection{Dataset Description}\label{ch:results:dataset}

We evaluate our methods on the ECG5000 dataset introduced in~\cite{chen_general_2015} and based on data from the BIDMC Congestive Heart Failure Database (chfdb)~\cite{baim_survival_1986,goldberger2000physiobank_short}. 
The record has 17\,998\,834 data points, containing 92\,584 heartbeats. The data was then pre-processed in two steps. Each heartbeat was extracted and then made equal length using interpolation.  Five classes are annotated, corresponding to the following labels: Normal (N), R-on-T Premature Ventricular Contraction (R-on-T PVC), Premature Ventricular Contraction (PVC), Supra-ventricular Premature or Ectopic beat (SP or EB), and unclassified beat (UB). Five thousand heartbeats were then extracted at random. The dataset was then split into a training set (500 data points) and a testing set (4500 data points) to report accuracy.

\subsection{Accuracy Results}

Table~\ref{tab:ecg5000scores} summarizes the accuracy and balanced accuracy scores of our proposed network and the comparison to related works. 
%  HERE WHAT DO WE COMPARE TO. 
The previously best performing network architecture on the ECG5000 dataset was the LSTM-FCN~\cite{karim_lstm_2018} which is a concatenation of an LSTM and a CNN network. We reproduced the results from the authors and further investigated the two branches of the network by splitting them, training them, and optimizing them as separate networks to fully evaluate and compare to the SoA. Four other network architectures from~\cite{du_attention-based_2018} are then included in the comparison in Table~\ref{tab:ecg5000scores} as well as a baseline of one nearest neighbor classifier.
%
%The TCN compared to the baseline proposed by the original authors of the selected dataset~\cite{chen_ucr_nodate}, our model is 1.7\% more accurate and excels by 34.1\% in balanced accuracy.
Our model is 1.7\% more accurate and excels by 34.1\% in balanced accuracy compared to the baseline proposed by the original authors of the selected dataset~\cite{chen_ucr_nodate}.

%
%Comparing the TCN to the baseline of one nearest neighbor classifier we see that the TCN has better accuracy results while also achieving much better balanced accuracy results (34.1\% better).
%
Our TCN has similar accuracy results as the previous SoA LSTM-FCN. By looking at the balanced accuracy score, we see that our TCN achieves much better results (16.5\% better) than the previous SoA and robustly classify the underrepresented classes in the dataset. 
Also, when considering the size and complexity of the models in Table~\ref{tab:ecg5000scores}, we observe that our TCN requires 27$\times$ fewer parameters (15\,k vs 405\,k) and 37$\times$ lower number of MACs (1\,MMACs vs 37\,MMACs), compared to the LSTM-FCN.  
Using only the LSTM reduces the number of parameters and MACs, but also yields to a lower accuracy (93.1\%) and balanced accuracy (68.9\%).

%Also, by looking at the LSTM network in Table~\ref{tab:ecg5000scores}, we see that the LSTM network has very comparable accuracy results but cannot correctly classify outlier classes that are not represented equally in the dataset.
%

% Talk somehow better about balanced accuracy (improvement in numbers preferably)

% Quantization talk 
Next, we consider the classification performance of the quantized TCN models, provided in Table~\ref{tab:results:summary}, and compare it to a quantized CNN+GRU that was trained on a different  dataset which distinguishes between five classes too~\cite{faraone2020aicas}. 
Quantizing the TCN weights and activations from float-32 to INT-8 representation yields a negligible accuracy loss of 0.2\% and 0.4\% using TFLite and NEMO, respectively. 
%
%This decrease in performance is minimal, and the resulting quantized models still perform at an exceptional level. 
Similarly, the work in~\cite{faraone2020aicas} has seen a 0.4\% accuracy loss after quantizing the CNN+GRU model form floating-point to INT-8 representation. The TCN quantized with NEMO has a 8.1\% higher accuracy than the quantized CNN+GRU.

\begin{table}[b!]
    \caption{Accuracy Scores of Methods on the ECG5000 Dataset}
    \label{tab:ecg5000scores}
    \vspace{-.2cm}
    \centering
    \resizebox{\linewidth}{!}{%
%    \begin{threeparttable}
    \begin{tabular}{c|ccrr}
 \textbf{Architecture} &  \textbf{Acc.~{[}\%{]}} &  \textbf{Bal. Acc.~{[}\%{]}}  & \textbf{\#Param.} & \textbf{MACs}\\ \hline
\rowcolor[HTML]{EFEFEF} 
TCN (ours) & \textbf{94.2} & \textbf{89.0} & \textbf{14\,883} & 1\,030\,260\\
LSTM-FCN~\cite{karim_lstm_2018}\tnote{1} & 94.1 & 72.5 & 404\,741 & 36\,982\,656 \\
CNN ~\cite{karim_lstm_2018}\tnote{2} & 93.4 & 81.5 & 266\,373 & 36\,844\,160 \\
\rowcolor[HTML]{EFEFEF} 
LSTM~\cite{karim_lstm_2018}\tnote{2} & 93.1 & 68.9 & 138\,373 & \textbf{138\,496}\\
1-NN (L2 dist.)~\cite{chen_ucr_nodate} & 92.5 & 54.9 & 70\,000 & 140\,000 \\
\rowcolor[HTML]{EFEFEF} 
AttLSTM-CNNs~\cite{du_attention-based_2018} & 85.4 & -  & - &- \\
Attention LSTM~\cite{du_attention-based_2018} & 85.3 & - &  - & -\\
\rowcolor[HTML]{EFEFEF} 
CNNs1D~\cite{du_attention-based_2018} & 84.8 & - & - &- \\
LSTM~\cite{du_attention-based_2018}  & 85.4 & -  & - & -
\end{tabular}
% \begin{tablenotes}\footnotesize
% %\item[1] Proposed method.
% \item[1] Network architecture trained and tested.
% \item[2] Network from paper, re-purposed and split up.
% \end{tablenotes}
%\end{threeparttable}
}
\end{table}

\begin{table*}[ht!]
\renewcommand{\arraystretch}{0.95}
  \centering
  \caption{Comparison Between TCN on GAP8, TCN on ARM Cortex M4F, and Related Work}\label{tab:results:summary}
  \vspace{-.2cm}
  {
    \footnotesize
    \begin{tabular}{lrrrrrr}
      \toprule
      Network & \multicolumn{5}{c}{TCN (ours)} & CNN+GRU~\cite{faraone2020aicas} \\
      \cmidrule(r){1-1} \cmidrule(r){2-6} \cmidrule(r){7-7}
      Platform & \multicolumn{2}{c}{GAP8} & \multicolumn{3}{c}{STM32L475} & nRF52832 \\
      \multirow{2}{*}{MCU} & \multicolumn{2}{c}{1+8$\times$CV32E40P} & \multicolumn{3}{c}{1$\times$Cortex M4F} & 1$\times$Cortex M4F \\
       & \multicolumn{2}{c}{@100\,MHz} & \multicolumn{3}{c}{@80\,MHz} & @64\,MHz\\
       \cmidrule(r){2-3} \cmidrule(r){4-6} \cmidrule(r){7-7}
      Deployment framework & NEMO/DORY & GAPflow & TFLite & \textsc{Cube.AI} TFLite & \textsc{Cube.AI} Keras & CMSIS-NN \\ 
      Dataset & ECG5000 & ECG5000 & ECG5000 & ECG5000 & ECG5000 & \cite{faraone2020aicas} \\
     \cmidrule(r){1-1} \cmidrule(r){2-2} \cmidrule(r){3-3} \cmidrule(r){4-4} \cmidrule(r){5-5} \cmidrule(r){6-6} \cmidrule(r){7-7}
      Input size & $1\times140$ & $1\times140$ & $1\times140$  & $1\times140$ & $1\times140$ & $1\times256$\\
      MACs & 1\,030\,260 & 2\,348\,925 & 2\,339\,994  & 2\,339\,994 & 2\,354\,114 & 3\,221\,244\\
      Memory (weights \& act.) & 26.63\,kB & 25.03\,kB & 35.86\,kB & 35.86\,kB & 113.40\,kB & 210.00\,kB \\
      \cmidrule(r){1-1} \cmidrule(r){2-2}\cmidrule(r){3-3} \cmidrule(r){4-4} \cmidrule(r){5-5} \cmidrule(r){6-6} \cmidrule(r){7-7}
      Full-prec. accuracy [\%] & 94.2 & 94.2 & 94.2 & 94.2 & 94.2 & 86.1 \\
      Quantized accuracy [\%]& 93.8  & 94.0 & 94.0 & 94.0 & not quantized & 85.7 \\
      \cmidrule(r){1-1} \cmidrule(r){2-2}\cmidrule(r){3-3} \cmidrule(r){4-4} \cmidrule(r){5-5} \cmidrule(r){6-6} \cmidrule(r){7-7}
      %Voltage [V]           & 0.8 & 1.2& ??\\
      Time/inference [ms]    &2.7 &20.2 & 188.0 & 126.5 & 374.3 &  94.8 \\
      Throughput [MMAC/s] & 381.58 &116.28 & 12.45 & 18.50 &  6.29 & 33.98 \\
      MACs/cycle & 3.795 &1.157 & 0.162 & 0.231 & 0.078 & 0.531 \\
      Power [mW] & 38.52 &41.67 & 45.54 & 42.75 & 47.68 & 20.65\\
      Energy/inference [mJ]   &0.10 &0.84 & 8.56 & 5.41 & 17.85 & 1.96\\
      En. eff. [GMAC/s/W] & 9.91 &2.79 & 0.27 & 0.43 & 0.13 &  1.64 \\
      \bottomrule
    \end{tabular}
  }
  
  \vspace{-.3cm}
\end{table*}

\subsection{Power and Energy}

\gapflow{
We perform power measurements to assess the energy consumption of our TCN implementations using five different toolchains on two \glspl{mcu}. Table~\ref{tab:results:summary} shows the comparison of the measurement between these methods; additionally, we include the power measurements from~\cite{faraone2020aicas}.
When looking at the ARM Cortex M4F implementation of the TCN, using a quantized TFLite model with \textsc{Cube.AI} provides the best performance. It has a speedup of around $1.5\times$ with respect to using TFLite barebone on the MCU.   
Both GAP8 implementations outperform the ARM ones, with NEMO and DORY tools, augmented by our optimized 1D kernels, consuming the least amount of energy per inference (0.10\,mJ) at highest energy efficiency (9.91\,GMAC/s/W), while also providing the highest performance (381.58\,MMAC/s) by executing one inference in only 2.7\,ms. The best GAP8 implementation is approximately $23.0\times$ more energy-efficient than the best deployment of the TCN onto the ARM Cortex M4F. Additionally, comparing to~\cite{faraone2020aicas}, it is around $6.0\times$ more energy-efficient than the CMSIS-NN method used to implement their network on the ARM Cortex M4F, while consuming 19.6$\times$ less energy.

}